\documentclass[10pt,a4paper]{article}
\usepackage[T1]{fontenc}
\usepackage[utf8]{inputenc}
\usepackage{lmodern}
\usepackage[margin=2.54cm]{geometry}
\usepackage{amsmath,amssymb}
\usepackage{graphicx}
\usepackage{booktabs,tabularx,array}
\usepackage{microtype}
\usepackage{tikz}
\usetikzlibrary{arrows.meta,positioning,calc,patterns}
\usepackage[numbers,sort&compress]{natbib}
\usepackage[hidelinks]{hyperref}
\usepackage{xcolor}
\definecolor{indigo}{HTML}{3F3B8C}
\definecolor{riskfill}{HTML}{FBEAE4}\definecolor{riskline}{HTML}{9A3412}
\definecolor{valfill}{HTML}{E3F4EC}\definecolor{valline}{HTML}{0B6B49}
\definecolor{feedback}{HTML}{6B3FA0}\definecolor{sand}{HTML}{F2EEE8}
\title{Which AI Use Cases Deserve a Pilot? A Risk and Value Framework for Funding, Scoping, and Scaling AI}
\author{Vishal Srivastava\\\small Whiting School of Engineering, Johns Hopkins University, Baltimore, MD 21218\\\small\texttt{vsrivas7@jhu.edu}\,\textperiodcentered\,ORCID \href{https://orcid.org/0000-0002-5651-8365}{0000-0002-5651-8365}}
\date{November 2026}
\begin{document}
\maketitle
\begin{abstract}
A composite score can rank AI use cases while leaving the permitted operating scope and the evidence required for expansion unspecified. We treat the unit of decision as a use case at a proposed scope (authority, human mediation, limits, volume, environment, population) and define three decisions: screen the scope, fund the experiment, and authorize expansion only when residual risk is acceptable and evidence of incremental net value applies at that scope. A risk track (R1--R4) is implemented as an explicit decision table with vetoes and documented precedence, and a value track follows four evidence stages with each transition needing its own evidence. Funding uses expected value of perfect information as an upper bound and the value of sample information for the specific experiment. A quantified expected-loss check, required where policy says so, computes the containment an expansion needs and credits a control only on evidence measured at the proposed scope; in a synthetic scenario the highest-value control design fails that test while a smaller scope passes. A reproducible Python package with 279 passing tests demonstrates the framework on synthetic refund, reconciliation and knowledge-assistant use cases, including a synthetic matched-scope fixture and a transferability stress test (value evidence is simulated, not realized), and compares it with a public AWS sample. Thresholds and appetite are prototype policy requiring field validation.
\end{abstract}

\section{Introduction}\label{sec:intro}
Organizations are asked to choose among many candidate AI projects. A composite score can rank AI use cases while leaving the permitted operating scope and the evidence required for expansion unspecified. This paper argues that the \emph{unit of decision} is an AI use case \emph{at a proposed operating scope}. Scope includes the system's authority (recommend, execute with approval, autonomous capped, autonomous uncapped), the human mediation around it (per-item, sampled or none), transaction limits, volume, environment and the affected population. A refund recommender that a human reads, an agent that executes each refund after per-item approval, and an agent that issues refunds autonomously up to a cap are the same business idea at three scopes with very different risk and evidence needs.

\begin{figure*}[t]
\centering
\resizebox{\textwidth}{!}{
\begin{tikzpicture}[
  font=\sffamily\scriptsize, >=Stealth, line cap=round,
  box/.style={draw=black!65, rounded corners=2.5pt, line width=0.6pt, align=center, inner sep=3pt},
  stp/.style={box, fill=indigo!7, draw=indigo!80!black, minimum width=3.02cm, text width=2.9cm, minimum height=6.3cm, anchor=north},
  ttl/.style={font=\sffamily\scriptsize\bfseries, text=indigo!80!black},
  rk/.style={box, draw=riskline, fill=riskfill, pattern=north east lines, pattern color=riskline!30, minimum height=0.46cm, minimum width=0.82cm, inner sep=1pt},
  vl/.style={box, draw=valline, fill=valfill, minimum height=0.46cm, minimum width=0.82cm, inner sep=1pt},
  outb/.style={box, fill=white, draw=black!60, text width=2.7cm},
  flow/.style={->, line width=0.9pt, draw=black!70},
  back/.style={->, line width=0.9pt, draw=feedback, dashed}
]
\def\w{3.02}\def\g{0.375}
\foreach \i/\t in {1/{Define the candidate},2/{Screen the pilot scope},3/{Choose and fund the experiment},4/{Run the pilot, collect evidence},5/{Decide whether to expand}}{
  \node[stp] (s\i) at ({(\i-1)*(\w+\g)+\w/2},0) {};
  \node[ttl, text width=2.8cm, align=center, anchor=north] at ([yshift=-3pt]s\i.north) {\i. \t};
}
\node[font=\sffamily\scriptsize, text width=2.7cm, align=center, anchor=north] at ([yshift=-1.0cm]s1.north)
  {Use case at a \textbf{proposed operating scope}:\\[2pt] authority $\cdot$ human mediation $\cdot$ transaction limit $\cdot$ volume $\cdot$ environment $\cdot$ affected population};
\node[font=\sffamily\scriptsize, text width=2.7cm, align=center, anchor=north] at ([yshift=-1.0cm]s2.north) {Prohibitions and mandatory controls first, then:};
\node[rk, anchor=north] (r1) at ([xshift=-0.93cm,yshift=-1.95cm]s2.north) {R1};
\node[rk, right=0.1cm of r1] (r2) {R2};
\node[rk, right=0.1cm of r2] (r3) {R3};
\node[font=\sffamily\scriptsize, text width=2.75cm, align=center, anchor=north] at ([yshift=-2.5cm]s2.north) {technical failures\\ deployment controls\\ business consequences};
\node[outb, anchor=south] at ([yshift=0.1cm]s2.south) {\textbf{Out:} maximum permitted pilot scope (may be smaller than the proposed expansion scope)};
\node[font=\sffamily\scriptsize, text width=2.7cm, align=center, anchor=north] at ([yshift=-1.0cm]s3.north) {Projected net value $\cdot$ feasibility $\cdot$ experiment cost $\cdot$ what this experiment can teach};
\node[vl, anchor=north] (v) at ([yshift=-2.55cm]s3.north) {V1--V4};
\node[font=\sffamily\scriptsize, text width=2.7cm, align=center, anchor=north] at ([yshift=-3.1cm]s3.north) {value hypotheses at intake (projected, not evidence)};
\node[outb, anchor=south] at ([yshift=0.1cm]s3.south) {\textbf{Out:} eligible experiment to fund and its measurement plan};
\node[font=\sffamily\scriptsize, text width=2.7cm, align=center, anchor=north] at ([yshift=-1.0cm]s4.north) {Projected value, untested controls and design assumptions meet actual pilot results};
\node[vl, anchor=north] at ([yshift=-2.55cm]s4.north) {E1--E4};
\node[font=\sffamily\scriptsize, text width=2.7cm, align=center, anchor=north] at ([yshift=-3.1cm]s4.north) {measured stage evidence, with the scope it covers};
\node[outb, anchor=south] at ([yshift=0.1cm]s4.south) {\textbf{Out:} control tests, error rates and effects, each tied to a scope};
\node[font=\sffamily\scriptsize, text width=2.7cm, align=center, anchor=north] at ([yshift=-1.0cm]s5.north) {Both at the \textbf{proposed expansion scope}:};
\node[rk, anchor=north, text width=2.5cm, minimum height=1.45cm] at ([yshift=-1.8cm]s5.north) {acceptable residual risk (R4); expected-loss check where policy requires it};
\node[vl, anchor=north, text width=2.5cm, minimum height=1.0cm] at ([yshift=-3.5cm]s5.north) {sufficient value evidence (E1--E4 stages)};
\node[outb, anchor=south] at ([yshift=0.1cm]s5.south) {\textbf{Out:} Scale $\cdot$ Extend pilot with constraints $\cdot$ Redesign $\cdot$ Stop};
\foreach \i/\j in {1/2,2/3,3/4,4/5}{ \draw[flow] (s\i.east) -- (s\j.west); }
\node[box, fill=white, draw=feedback, line width=0.9pt, minimum width=16.6cm, text width=16.2cm, anchor=north] (app) at (8.3,-6.9)
  {\textbf{Evidence applicability} links the steps: every control test, error measurement and value estimate is classed as\\
   \textbf{applies as measured} $\cdot$ \textbf{needs a stated, evaluated transfer} $\cdot$ \textbf{must be remeasured} $\cdot$ \textbf{missing or conflicting}};
\foreach \i in {1,...,5}{ \draw[<->, line width=0.8pt, draw=feedback] (s\i.south) -- (s\i.south |- app.north); }
\node[box, fill=feedback!10, draw=feedback, minimum width=16.6cm, text width=16.2cm, anchor=south] (fb) at (8.3,0.85)
  {\textbf{Scope or evidence changes} (authority, mediation, limit, volume, population; a new blocker, test failure or finding) $\rightarrow$ revisit the affected decisions: screening (2), funding (3) and expansion (5)};
\foreach \i in {2,3,5}{ \draw[back] (fb.south -| s\i.north) -- (s\i.north); }
\end{tikzpicture}}
\caption{Decision framework for one candidate (a use case at a proposed operating scope). Screening, funding and expansion are the three decisions; the pilot between funding and expansion produces the evidence that expansion needs. R1--R3 are technical failures, deployment controls and business consequences, R4 is residual risk, V1--V4 are value stages and E1--E4 the evidence levels that support them. Both the risk gate (residual risk, with the expected-loss check where policy requires it) and the value gate must pass at the proposed expansion scope. Evidence applicability links all steps, and a change of scope or evidence (dashed) sends the candidate back to the affected decisions. R-BED and V-BED labels follow a presentation slide; table values, thresholds and appetite are PilotGate prototype policy.}\label{fig:framework}
\end{figure*}
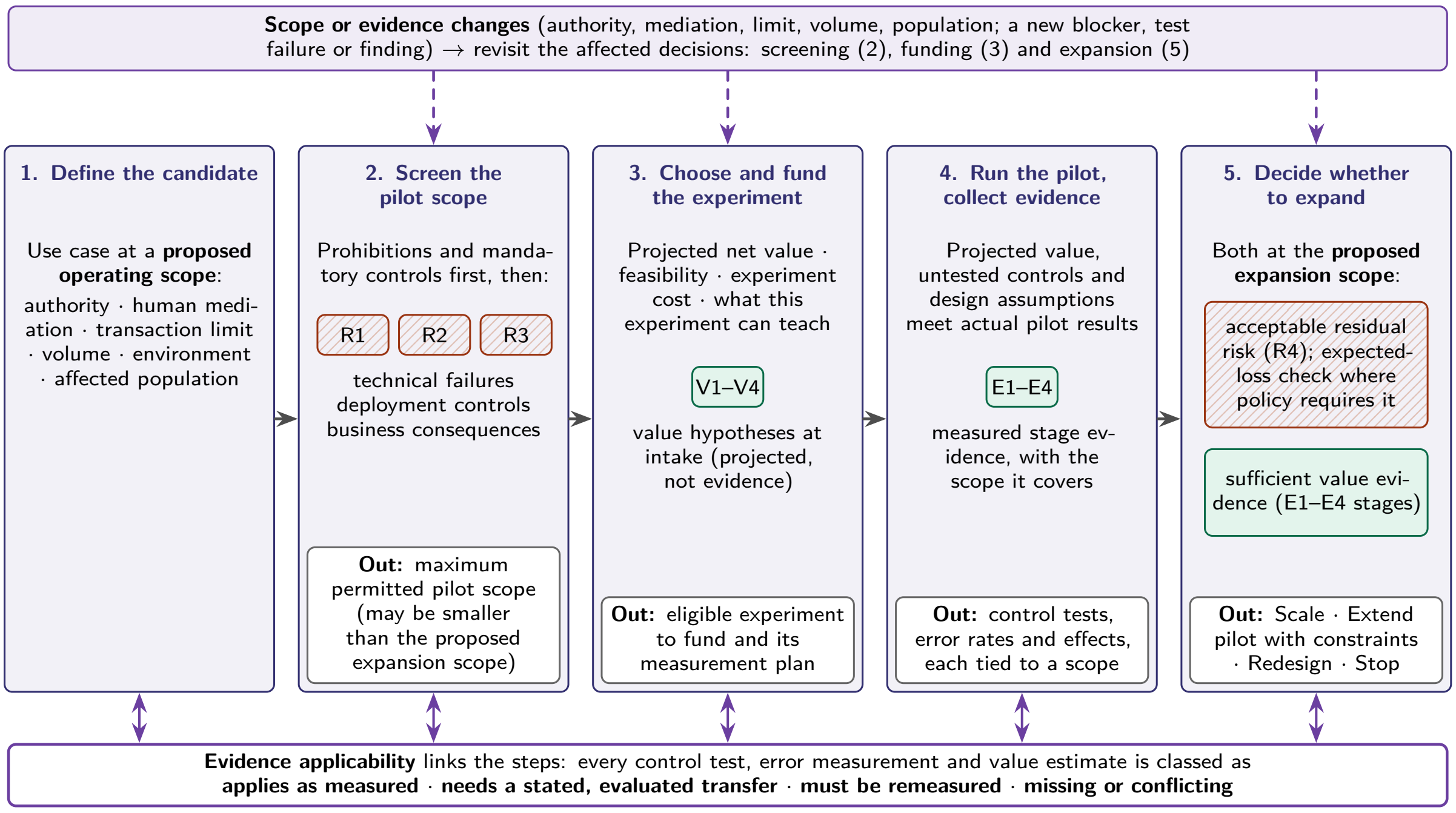

Three decisions follow, shown in Figure~\ref{fig:framework}. \textbf{Screen the scope}: what the system may do during the pilot and under which constraints. \textbf{Fund the experiment}: whether an eligible pilot justifies its cost given projected value, feasibility and what that specific experiment can teach. \textbf{Authorize expansion}: whether both the residual risk is acceptable \emph{and} there is sufficient evidence of incremental net value at the proposed scope. Whenever scope changes, the framework asks which evidence still applies. This is the conceptual contribution. The comparison with a public AWS sample (Section~\ref{sec:aws}) is an evaluation example only.

\paragraph{Worked example in brief.} The refund candidates in Table~\ref{tab:example} are synthetic. RA, the recommender, can run a live pilot with human-approved action and, with evidence at the evaluated scope, passes both gates for expansion (incremental net value about \$465,234 per year, interval \$220,294 to \$710,174). RB, the approval-gated executor, cannot clear the risk gate at its proposed target, and the evidence from its pilot does not carry to the larger scope. Our recommendation is to redesign the scope and then reassess pilot funding, because no EVSI was calculated for a redesigned lower scope. RC, the autonomous executor, may only run in shadow mode, and its shadow run is funded for the error-rate information it yields (a risk decision about the next pilot rung) rather than for value. We carry these three through the paper.

\paragraph{Contributions.} (1) A scope-aware framework with three decisions and an explicit evidence-applicability rule. (2) A risk decision table (R1--R4) with vetoes, scope-specific control requirements and documented precedence, kept separate from quantified loss. (3) A stage-linked value track (V1--V4, E1--E4) in which every transition needs its own evidence, projections are hypotheses, and decision influence is a metric only where value depends on changing decisions. (4) A funding rule using EVPI as an upper bound and EVSI for the designed experiment. (5) A quantified expected-loss check inside the expansion decision (Section~\ref{sec:quant}): required containment, a capacity ceiling, an evidence standard under which a control is credited only on scope-applicable live evidence, and a comparison of control designs. (6) A reproducible Python implementation with 279 passing tests, and synthetic demonstrations, including a synthetic matched-scope fixture and a transferability stress test (value evidence is simulated, not realized).

\section{Framework}\label{sec:framework}
\subsection{Unit of decision and scope}
A candidate is a tuple of use case and proposed scope. Authority is ordered RECOMMEND $<$ EXECUTE\_WITH\_APPROVAL $<$ AUTONOMOUS\_CAPPED $<$ AUTONOMOUS\_UNCAPPED. A transaction limit is capped, uncapped, not\_applicable or unknown. Higher authority and weaker mediation never inherit evidence gathered at a lower authority or stronger mediation; limit and volume may inherit it only through a stated, justified, evaluated transfer assumption within policy caps.

\subsection{Decision 1: screen the scope}\label{sec:risk}
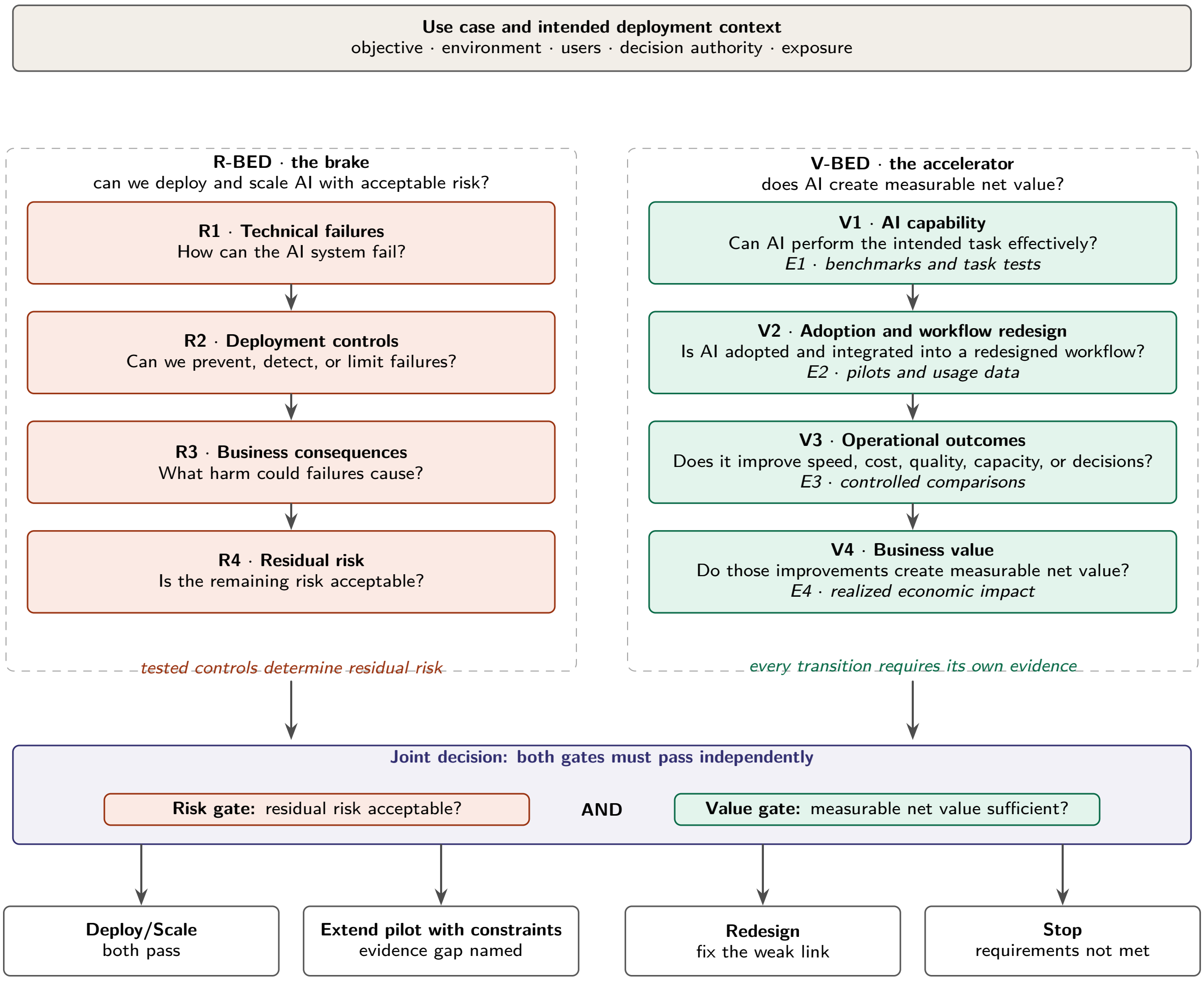
\begin{figure*}[t]
\centering
\resizebox{\textwidth}{!}{
\begin{tikzpicture}[
  font=\sffamily\scriptsize, >=Stealth, line cap=round,
  box/.style={draw=black!65, rounded corners=2.5pt, line width=0.6pt, align=center, inner sep=3pt},
  rb/.style={box, draw=riskline, fill=riskfill, text width=7.0cm, minimum height=1.12cm},
  vb/.style={box, draw=valline, fill=valfill, text width=7.0cm, minimum height=1.12cm},
  flow/.style={->, line width=0.8pt, draw=black!70},
  col/.style={draw=black!35, dashed, rounded corners=3pt, minimum width=7.8cm, minimum height=7.15cm, anchor=north}
]
\node[box, fill=sand, text width=15.9cm, minimum height=0.9cm] (ctx) at (8.3,0) {\textbf{Use case and intended deployment context}\\ objective $\cdot$ environment $\cdot$ users $\cdot$ decision authority $\cdot$ exposure};
\node[col] (cl) at (4.05,-1.5) {};
\node[col] (cr) at (12.55,-1.5) {};
\node[font=\sffamily\scriptsize, align=center, text width=7.5cm] at ([yshift=-0.35cm]cl.north) {\textbf{R-BED $\cdot$ the brake}\\ can we deploy and scale AI with acceptable risk?};
\node[font=\sffamily\scriptsize, align=center, text width=7.5cm] at ([yshift=-0.35cm]cr.north) {\textbf{V-BED $\cdot$ the accelerator}\\ does AI create measurable net value?};
\foreach \i/\t/\q in {1/{Technical failures}/{How can the AI system fail?},2/{Deployment controls}/{Can we prevent, detect, or limit failures?},3/{Business consequences}/{What harm could failures cause?},4/{Residual risk}/{Is the remaining risk acceptable?}}{
  \node[rb] (r\i) at (4.05,{-2.8-(\i-1)*1.5}) {\textbf{R\i\ $\cdot$ \t}\\ \q};
}
\foreach \i/\t/\q/\e in {1/{AI capability}/{Can AI perform the intended task effectively?}/{E1 $\cdot$ benchmarks and task tests},2/{Adoption and workflow redesign}/{Is AI adopted and integrated into a redesigned workflow?}/{E2 $\cdot$ pilots and usage data},3/{Operational outcomes}/{Does it improve speed, cost, quality, capacity, or decisions?}/{E3 $\cdot$ controlled comparisons},4/{Business value}/{Do those improvements create measurable net value?}/{E4 $\cdot$ realized economic impact}}{
  \node[vb] (v\i) at (12.55,{-2.8-(\i-1)*1.5}) {\textbf{V\i\ $\cdot$ \t}\\ \q\\ \emph{\e}};
}
\foreach \a/\b in {1/2,2/3,3/4}{\draw[flow] (r\a.south)--(r\b.north); \draw[flow] (v\a.south)--(v\b.north);}
\node[font=\sffamily\scriptsize\itshape, text=riskline] at (4.05,-8.6) {tested controls determine residual risk};
\node[font=\sffamily\scriptsize\itshape, text=valline] at (12.55,-8.6) {every transition requires its own evidence};
\node[box, fill=indigo!7, draw=indigo!80!black, text width=15.9cm, minimum height=1.35cm] (joint) at (8.3,-10.35) {};
\node[font=\sffamily\scriptsize\bfseries, text=indigo!80!black] at (8.3,-9.85) {Joint decision: both gates must pass independently};
\node[box, draw=riskline, fill=riskfill, text width=5.6cm] (rg) at (4.4,-10.55) {\textbf{Risk gate:} residual risk acceptable?};
\node[font=\sffamily\scriptsize\bfseries] at (8.3,-10.55) {AND};
\node[box, draw=valline, fill=valfill, text width=5.6cm] (vg) at (12.2,-10.55) {\textbf{Value gate:} measurable net value sufficient?};
\draw[flow] (4.05,-8.8)--(4.05,-9.6); \draw[flow] (12.55,-8.8)--(12.55,-9.6);
\foreach \n/\x/\t/\s in {1/2.0/{Deploy/Scale}/{both pass},2/6.1/{Extend pilot with constraints}/{evidence gap named},3/10.5/{Redesign}/{fix the weak link},4/14.6/{Stop}/{requirements not met}}{
  \node[box, fill=white, text width=3.55cm, minimum height=0.95cm] (o\n) at (\x,-12.35) {\textbf{\t}\\ \s};
  \draw[flow] (joint.south -| o\n.north)--(o\n.north);
}
\end{tikzpicture}}
\caption{Presentation-level structure of the R-BED (risk) and V-BED (value) frameworks, redrawn from the slide supplied by the author. R-BED and V-BED are proposed frameworks of V.~Srivastava; the figure shows their structure only and is not a result of this paper, which adds operating scope, evidence applicability and the quantified check (Figure~\ref{fig:framework}). The slide's third outcome, ``Pilot, with constraints'', is labelled here with the PilotGate action ``Extend pilot with constraints'', which continues an existing pilot; the pilot-screen permission is ``Start constrained pilot''.}\label{fig:rbed}
\end{figure*}

Screening follows the risk track presented as R1 technical failures, R2 deployment controls, R3 business consequences and R4 residual risk (Figure~\ref{fig:rbed}). We implement it as a decision table in which tested controls lower the \emph{residual} assessment relative to the inherent one. R1 is the failure level, from an ordinal judgment or, when a measured 95\% upper bound on the error rate covers the scope, from fixed cut points. R2 is the maximum of human-control, exposure and monitoring levels. R3 is the maximum of severity, propagation and recovery levels. Table~\ref{tab:risktable} maps R2 and R3 to a base tier, which R1 adjusts by one step. Hard blockers prohibit the proposed action; vetoes force the RESTRICTED tier before any other rule. An appetite tier per authority then decides whether the residual rating is acceptable (prototype policy: pilot HIGH; scale HIGH, HIGH, MEDIUM, NONE across the four authority levels). Controls count only if tested at the proposed scope (Section~\ref{sec:applic}). Three ratings are reported (inherent materiality, inherent before controls, residual) and ordinal judgments are never multiplied or mixed with quantified loss, which is reported on a separate loss view.

\begin{table}[t]
\centering
\footnotesize
\caption{PilotGate base risk tier from deployment-control level (R2, rows) and business-consequence level (R3, columns). This table is prototype policy, not a published R-BED table. R1 (technical failure) moves the tier by at most one step; any veto yields RESTRICTED first.}
\label{tab:risktable}
\begin{tabularx}{\textwidth}{lXXX}
\toprule
\textbf{R2 control level} & \textbf{R3 = 1} & \textbf{R3 = 2} & \textbf{R3 = 3} \\
\midrule
1 & LOW & LOW & MEDIUM \\
2 & LOW & MEDIUM & HIGH \\
3 & MEDIUM & HIGH & CRITICAL \\
\bottomrule
\end{tabularx}
\end{table}

\subsection{Decision 2: fund the experiment}
Eligibility comes from decision 1. Among eligible pilots, funding uses the intake value distribution, feasibility and information value. EVPI, the expected value of perfect information about the expansion decision, is only an upper bound. For each recommendation we also state what the \emph{specific} experiment can learn: a normal--normal preposterior gives EVSI for a designed pilot of $n$ cases per arm over a three-year horizon with a stated per-case standard deviation. A shadow run observes errors but not value, so its EVSI on value is zero; it may be funded for risk information or an independent rationale, which we name. An EVSI figure is stated with the experiment it describes: it may be that of a hypothetical design the screen does not permit, or of a scale decision the structural risk rule cannot let reach SCALE at the proposed scope; neither is a funding justification for that scope. If the information case is not supported (EVSI below cost) and no independent rationale is stated, the pilot is not recommended.

\subsection{Decision 3: authorize expansion}
Both gates must pass at the proposed scope. The \emph{risk gate} uses the R4 residual tier against appetite with control tests applicable to the proposed scope. The \emph{value gate} uses the stage-linked evidence ladder below.

\paragraph{Value track.} Following the presentation slide, V1/E1 comprises benchmarks and task tests, V2/E2 pilots and usage data, V3/E3 controlled comparisons, and V4/E4 realized economic impact. Every transition needs its own evidence, so a level is the highest consecutive stage with measured evidence; an intake projection is a hypothesis (E0). A controlled improvement in an operational outcome can support E3; an annualized estimate still depends on costing and deployment assumptions; E4 requires realized economic impact. A real E4 record needs observed operating costs (invoices or metered usage) and documented savings or redeployed capacity at the scope where they occurred. In this paper E4 is \emph{simulated}: the run cost is an assumed 105\% of the intake estimate, so the dollars verify the logic and are not realized impact. Annualized figures at the proposed scope are forecasts. We treat durability and transferability separately: durability uses configurable stability thresholds (a minimum study size, duration and stability share, which are policy assumptions and not attributed to any source), and transferability is handled by evidence applicability.

\paragraph{Mechanisms and decision influence.} Each value claim names a mechanism (decision change, speed, cost, quality or capacity), a baseline, an observable outcome and costing assumptions. Decision influence is measured only for use cases whose value depends on changing decisions. For others (for example approval-gated refunds where the approval decision is unchanged) speed, cost, quality or capacity is measured, so unchanged decisions can coexist with positive value. States are applicable, not\_applicable and unknown, and are tested separately. Strong capability and adoption can still yield zero or negative incremental value (K1 below).

\begin{table}[t]
\centering
\footnotesize
\caption{Stage evidence reached by each synthetic use case (E4 economics are simulated). The level is the highest consecutive measured stage; a projection alone is E0. Durability is reported separately (policy thresholds).}
\label{tab:stages}
\begin{tabularx}{\textwidth}{lllllll}
\toprule
\textbf{Use case} & \textbf{V1/E1} & \textbf{V2/E2} & \textbf{V3/E3} & \textbf{V4/E4} & \textbf{Level} & \textbf{Durable} \\
\midrule
RA & measured & measured & measured & measured & E4 & yes \\
RB & measured & measured & measured & measured & E4 & yes \\
K1 & measured & measured & measured & measured & E4 & yes \\
P1 & measured & measured & weak & none & E2 & no \\
RC & measured & none & none & none & E1 & no \\
\bottomrule
\end{tabularx}
\end{table}

\begin{table}[t]
\centering
\footnotesize
\caption{Value mechanisms. Decision influence is a metric only where value depends on changing a decision (applicable); otherwise it is not\_applicable, and an unrecorded state is unknown. RB's refund approval is unchanged by the tool, yet speed still creates value.}
\label{tab:mech}
\begin{tabularx}{\textwidth}{lllX}
\toprule
\textbf{Use case} & \textbf{Mechanism} & \textbf{Decision influence} & \textbf{Observable outcome} \\
\midrule
RA & decision\_\allowbreak{}change & applicable & Cost per eligible case: handling minutes, leakage from wrong refunds, rework. \\
RB & speed & not\_\allowbreak{}applicable & Handling minutes per case after the approval decision, net of approver minutes. \\
RC & capacity & not\_\allowbreak{}applicable & Share of refunds completed without staff time; errors and recovery cost. \\
P1 & speed & not\_\allowbreak{}applicable & Minutes to complete an unchanged reconciliation, before versus after. \\
K1 & decision\_\allowbreak{}change & applicable & Cost per contact: handling minutes, errors, rework; plus CSAT as guardrail. \\
D1 & decision\_\allowbreak{}change & unknown & Cost per dispute. \\
\bottomrule
\end{tabularx}
\end{table}

\subsection{Evidence applicability}\label{sec:applic}
For each control test, error measurement and value estimate, the scale decision asks whether it applies as measured, needs a stated, justified, evaluated transfer, or must be remeasured. Authority and mediation never transport upward. A control-test failure newer than, or unordered against, a pass revokes the credit; legacy or undated records get no credit at scale; test age is capped at 180 days at the as-of date (policy). Volume transfers only within a ratio cap of $6\times$ and limits only within a ratio of $1.0$, with an effect haircut.

\subsection{Joint decision and executive actions}
Precedence is: a prohibition gives BLOCKED; a scope contradiction or veto gives REDESIGN; an incomplete proposal gives DEFER; contradictory value evidence gives EXTEND\_PILOT; value FAIL gives REDESIGN or STOP; risk FAIL gives REDESIGN; both PASS gives SCALE; otherwise EXTEND\_PILOT. Executives see four actions: Deploy/Scale, Extend pilot with constraints, Redesign and Stop. "Extend pilot with constraints" continues an \emph{existing} pilot; permission to \emph{start} a constrained pilot is the separate outcome of the pilot screen. BLOCKED and DEFER are implementation statuses. A blocker \emph{prohibits the proposed action} and maps to Stop; because it covers every rung of the ladder, no lower-scope experiment is available. A deferral means evidence or scope information is missing or unresolved; it maps to no action, and a lower-scope experiment is allowed only if that lower scope is itself complete and passes the pilot screen. Table~\ref{tab:exec} gives the mapping for the example.

\section{Quantifying control investment}\label{sec:quant}
Ordinal tiers say how risky a scope is, not whether a control is large enough. Where policy requires it, the expansion decision therefore adds a \emph{quantified expected-loss check} at the proposed scope. The check is an additional risk check: it never replaces the ordinal rating, the vetoes or the mandatory controls, and a pass satisfies that check only. The value gate stays independent, so no projected value offsets a failed check, and a pass never rescues a value failure. If the check is required and its inputs or applicable evidence are missing, the result is an explicit INSUFFICIENT (the decision is then EXTEND\_PILOT), never a silent pass. If policy does not require the check, it is reported for information and no decision changes.

\subsection{Expected loss, required containment and capacity}
Let $V$ be eligible actions per day and $N = 365\,V$ per year (one-year horizon, no discounting). Let $q$ be the share of actions that are unsafe attempts, $l$ the loss per escape and $d$ the share of unsafe attempts contained before execution. Expected annual loss is $L = N\,q\,(1-d)\,l$, computed separately for ordinary execution error and strategic abuse (mutually exclusive modes). Against an annual appetite $A$ the required containment is $d_{\min} = \max\{0,\,1 - A/(N q l)\}$. Containment cannot exceed what the control can deliver. For uniformly sampled review with recall $\rho$ and capacity $B$ reviews per day, $d(V) = \rho\min\{1, B/V\}$, so the modeled ceiling on volume is $V_{\max} = \rho B + A/(365\,q\,l)$ above $B$ (Appendix~\ref{app:deriv}). Mandatory per-item approval behaves differently: above capacity the queue grows by $V-B$ per day and throughput is limited to $B$; lower coverage is modeled only if the policy explicitly permits unreviewed overflow. A post-payment audit earns no pre-payment containment, because the payment has already settled. If it recovers funds, the recovery lowers the loss in a separate term and is never counted as containment.

Deterrence is optional and applies only where deliberate abuse is modeled. With abuse payoff $G$, penalty $C$ and honest payoff $T$, a risk-neutral actor abuses only if detection is below $p^* = (G-T)/(G+C)$, the break-even of an expected-payoff comparison of the kind used in the economic model of crime~\cite{becker1968crime}. The threshold follows directly from the stated payoffs. Abuse pays $G$ if undetected and costs the penalty $C$ if detected, so its expected payoff at detection probability $p$ is $(1-p)G - pC$, and the actor abuses only if this exceeds the honest payoff $T$, that is, only if $p < (G-T)/(G+C)$. Detection is not containment: a detected abuse can still have been paid. Equality at $p^*$ is indifference and is not credited as deterrence. In a separate illustration ($G=2$, $C=3$, $T=0.3$, so $p^*=0.34$) audit detection $0.85\cdot 600/V$ equals $p^*$ at exactly 1,500 cases per day (indifference), strictly exceeds it only below 1,500 and is 31\% at 1,644.

\subsection{What the required gate accepts as evidence}\label{sec:evidence}
A synthetic scenario input (a design recall, a flag rate, a recovery rate, an assumed detection probability) is a planning value. It drives the figures, the design comparison and the modeled boundaries, but \emph{the required gate never credits a planning value by itself}. Every credit that can lower the expected loss is
\[ \text{credited value} = \min\{\text{design value},\ \underline{p}\}\times r, \]
where $\underline{p}$ is the exact one-sided lower confidence bound (level 0.05) on a proportion measured by observed, live seeded attempts whose scope covers the proposed scope, and $r\le 1$ is the stated risk-side ratio when the evidence is carried to the proposed scope by a transfer assumption. The rule applies to reviewer or control recall, to pipeline containment, to the detection recall behind a post-payment audit, to recovery and to deterrence. A pipeline is credited only on a representative end-to-end test, or on stage records in which the second stage was measured \emph{among the attempts that escaped the first}; a second-stage rate measured on all attempts earns nothing. Deterrence is credited only if the lower bound on detection strictly exceeds $p^*$ and the payoffs $G$, $C$, $T$ are a stated, justified and evaluated assumption, since seeded attempts cannot measure them. Self-reported, non-live or inapplicable evidence earns no credit.

Transfer needs its own rule. A transfer record carries a value-side effect ratio for the savings effect, and that ratio is not a recall ratio: it is never applied to, or substituted for, one. Coverage alone, through a transfer assumption that states no risk-side recall ratio, earns no control credit (the check is INSUFFICIENT). A stated ratio, such as 0.85, multiplies the credited recall; 1.0 is an explicit, evaluated claim of no haircut.

The outcome has three branches. \textbf{Fail} if the stated design fails the appetite on its own, or if even the optimistic reading of the evidence, $\min\{\text{design},\ \text{exact upper bound}\}$, fails (for example 0 of 59 caught). \textbf{Pass} if the credited loss is within appetite. Otherwise \textbf{insufficient}: the design would pass but the evidence cannot yet show that it delivers what the design assumes, and the output states how many seeded attempts would be needed. The attempt rate $q$ and the loss per escape $l$ remain stated inputs that this gate does not test; they are reported, with stress scenarios that never gate. Appendix~\ref{app:evidence} gives the status codes and the three bypasses of an earlier version that this standard closes.

\subsection{Worked scenario}
\begin{figure*}[t]
\centering
\includegraphics[width=\textwidth]{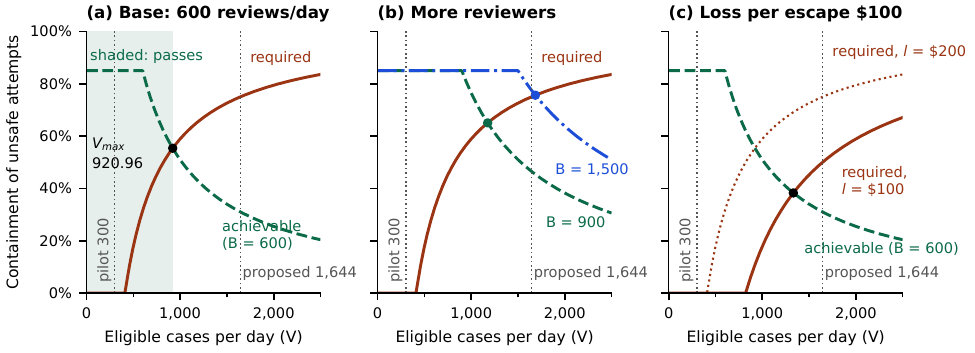}
\caption{Required and achievable containment against eligible cases per day for the synthetic scenario RF (1\% unsafe attempts, \$200 per escape, \$300,000 per year appetite, 85\% recall, 365 days, one-year horizon). Solid: required containment $d_{\min}(V)$; dashed: containment achievable by uniformly sampled review. Shading in (a): volumes at which the modeled expected-loss test passes ($V\le 920.96$). (b) More reviewers (ceilings 1,175.96 and 1,685.96). (c) A lower loss per escape, only if the effect of a limit on loss is defensibly estimated. A pass here is necessary, not sufficient: scale authorization still needs the ordinal risk, evidence, mandatory-control and value conditions.}\label{fig:contain}
\end{figure*}

The synthetic scenario RF has 1\% unsafe attempts, \$200 per escape, an appetite of \$300,000 per year (\$821.92 per day), 600 reviews per day at 85\% recall, and 300 cases per day in the pilot against 1,644 proposed (Table~\ref{tab:qworked}, Figure~\ref{fig:contain}). At 300 cases per day the expected loss with no control is \$219,000, below the appetite, so no containment is required. This is the teaching point: a pass at pilot volume does not show that the control can meet the expansion requirement. At 1,644 cases per day about 75\% containment is required while the current design achieves about 31\%. The modeled ceiling is 920.96 cases per day: 920.9 passes, the ceiling itself passes (equality) and 921.0 fails. At 1,644 cases per day the minimum capacity is 1,450.64 reviews per day, so 1,451 is the smallest whole number, and 1,450 fails. These boundaries hold under the stated point assumptions; whether the control meets them on evidence is the question of Section~\ref{sec:evidence}.

\begin{table}[t]
\centering
\footnotesize
\caption{Worked scenario RF (synthetic): 1\% unsafe attempts, \$200 per escape, \$300,000 per year appetite, 600 reviews per day at 85\% recall. At the pilot volume no containment is required; at the proposed volume about 75\% is required and sampled review achieves about 31\%.}
\label{tab:qworked}
\begin{tabularx}{\textwidth}{lXXXXX}
\toprule
\textbf{Cases/day (V)} & \textbf{Actions/yr (N = 365 V)} & \textbf{Expected loss, no control} & \textbf{Required containment} & \textbf{Achieved, sampled review} & \textbf{Expected loss, sampled review} \\
\midrule
300 & 109,500 & \$219,000 & 0.0\% & 85.0\% & \$32,850 \\
1,644 & 600,060 & \$1,200,120 & 75.0\% & 31.0\% & \$827,820 \\
\bottomrule
\end{tabularx}
\end{table}

Table~\ref{tab:qengine} runs the same cases through the decision path. Pilot-scope evidence does not apply at 1,644 cases per day without a stated and evaluated transfer, and when the transfer record states only a value-side effect ratio the check is INSUFFICIENT. With matched-scope evidence the engine reproduces the boundaries: 1,450 reviews per day fail and 1,451 pass. An observed 0 of 59 fails even at 1,451 reviews per day, because the design's 85\% recall is credited only through evidence. Mandatory per-item approval at 600 cases per day without permitted overflow fails as a queue.

\begin{table}[t]
\centering
\scriptsize
\caption{Engine decisions with the quantified check required. Containment is credited only on scope-applicable live evidence (Section~\ref{sec:evidence}); ``matched'' evidence means seeded attempts measured at the proposed scope. The expected loss shown is on the evidence-credited basis. The decision combines the check with RA's other (synthetic) risk and value evidence, held fixed.}
\label{tab:qengine}
\begin{tabularx}{\textwidth}{XXll}
\toprule
\textbf{Case (RA at 1,644 cases/day; quantified check required)} & \textbf{Quantified check} & \textbf{Expected loss, \$/yr (credited)} & \textbf{Decision} \\
\midrule
no quantified inputs & insufficient (no inputs) & -- & Extend pilot \\
evidence at pilot scope, no transport & insufficient (evidence not applicable) & -- & Extend pilot \\
pilot-scope evidence, value-only transport record, 1,451 reviews/day & insufficient (no risk-side rule) & -- & Extend pilot \\
pilot-scope evidence, risk-side recall ratio 0.85, 1,451 reviews/day & insufficient (evidence too weak) & \$434,826 & Extend pilot \\
sampled review, 600 reviews/day & fail & \$827,820 & Redesign \\
sampled review, 1,450 reviews/day & fail & \$300,395 & Redesign \\
sampled review, 1,451 reviews/day & pass & \$299,774 & Scale \\
sampled review, 1,451 reviews/day, observed 0/59 caught & fail & \$1,200,120 & Redesign \\
pipeline d1 = 1, no evidence & insufficient (evidence not applicable) & -- & Extend pilot \\
pipeline d = 0.80, representative end-to-end 59/59 & pass & \$240,024 & Scale \\
mandatory per-item, 600/day, no overflow & fail (queue) & \$180,018 & Redesign \\
mandatory per-item, 600/day, overflow permitted & fail & \$827,820 & Redesign \\
post-payment audit, 50\% recovery (recovery evidence 59/59) & fail & \$1,013,970 & Redesign \\
\bottomrule
\end{tabularx}
\end{table}

\subsection{Comparing control designs}
\begin{figure}[t]
\centering
\includegraphics[width=\textwidth]{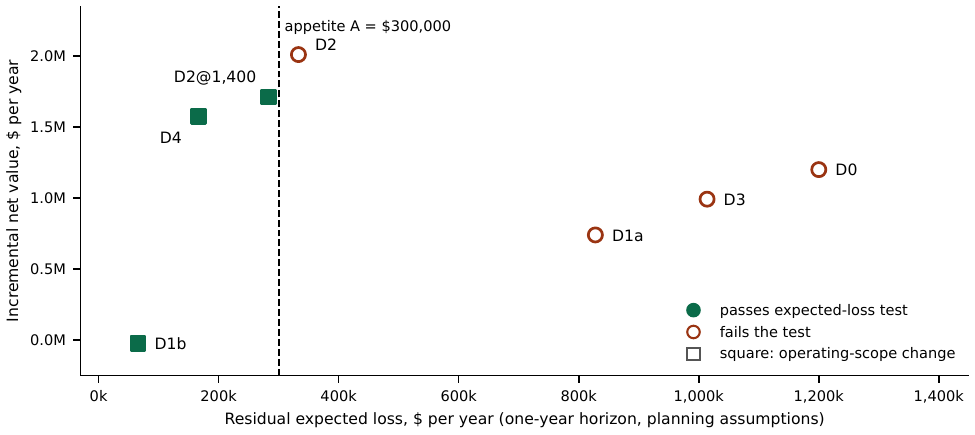}
\caption{Residual expected loss and incremental net value of the compared control designs at the proposed scope (synthetic RF, one year, manual-process baseline, avoided loss counted once). The dashed line is the \$300,000 appetite. Filled marks pass the expected-loss test applied to the \emph{stated design assumptions}; whether a design earns a pass on evidence is in Table~\ref{tab:qevpass}. Squares change the operating scope (served volume, volume or limit). D2 passes at automated-check recall of at least 0.8824; the \$100-limit result assumes a loss per escape of \$100; the recovery rate, the share of savings kept on reviewed cases and the cap's effect are planning assumptions.}\label{fig:designs}
\end{figure}

Table~\ref{tab:qdesigns} compares six designs with one baseline (the current manual process) and one horizon (one year), counting avoided loss once. A change in volume or in the transaction limit is a change of operating scope and is labelled as such. D2, automated pre-payment checks with escalation, has the highest net value (\$2,009,787 per year) and fails the expected-loss test: its residual expected loss is \$333,033, 11\% above the appetite. The failure is narrow. The exact break-even automated-check recall is $d_{\min}/\rho = 0.8824$: the expected loss is \$300,102 just below it and \$299,898 just above. Restricting volume to 1,400 cases per day (expected loss \$283,605, net value \$1,711,497) or adding a \$100 limit (\$166,517, \$1,576,244, with the lower loss per escape an assumption) passes. Figure~\ref{fig:designs} shows that the highest-value design is on the wrong side of the risk line, which the table alone does not.

\begin{table}[t]
\centering
\scriptsize
\caption{Design comparison (synthetic RF; manual-process baseline; one-year horizon; avoided loss counted once, inside net value). D0 no added control; D1a per-item approval with unreviewed overflow; D1b per-item approval without overflow (served volume limited to 600/day); D2 automated checks with escalation; D3 post-payment audit (recovery 50\% modeled separately, no pre-payment containment); D4 D2 with a \$100 limit. ``Planning test'' applies the \$300,000 appetite to the stated design assumptions; ``matched evidence'' is the same test with containment credited on 59 live seeded attempts, none missed, measured at that scope.}
\label{tab:qdesigns}
\begin{tabularx}{\textwidth}{lX>{\raggedright\arraybackslash}p{1.0cm}>{\raggedright\arraybackslash}p{1.65cm}>{\raggedright\arraybackslash}p{1.4cm}>{\raggedright\arraybackslash}p{1.3cm}>{\raggedright\arraybackslash}p{1.65cm}>{\raggedright\arraybackslash}p{1.3cm}>{\raggedright\arraybackslash}p{1.3cm}}
\toprule
\textbf{Design} & \textbf{Scope} & \textbf{d} & \textbf{Residual expected loss} & \textbf{Control cost} & \textbf{False blocks/yr} & \textbf{Net value} & \textbf{Planning test} & \textbf{Matched evidence} \\
\midrule
D0 & proposed scope & 0.000 & \$1,200,120 & \$0 & 0 & \$1,200,120 & fail & fail \\
D1a & proposed scope & 0.310 & \$827,820 & \$394,200 & 0 & \$740,220 & fail & fail \\
D1b & scope change: served volume restricted to B & 0.850 & \$65,700 & \$394,200 & 0 & -\$21,900 & queue & pass \\
D2 & proposed scope & 0.723 & \$333,033 & \$42,568 & 594 & \$2,009,787 & fail & fail \\
D3 & proposed scope & 0.000 & \$1,013,970 & \$394,200 & 0 & \$992,070 & fail & fail \\
D4 & scope change: limit \$100 & 0.723 & \$166,517 & \$42,568 & 594 & \$1,576,244 & pass & pass \\
D2 & scope change: volume 1,400/day & 0.723 & \$283,605 & \$36,250 & 506 & \$1,711,497 & pass & pass \\
\bottomrule
\end{tabularx}
\end{table}

\paragraph{Sensitivity.} Varying recovery from 0 to 75\% and loss per escape over \$100, \$200 and \$300, D2 remains the highest-value design in every cell and a post-payment audit is never feasible; recovery moves only D3. The conclusion that the highest-value design fails the expected-loss test holds at \$200 and \$300 per escape and does not hold at \$100, where D2 passes. These are sensitivity ranges, not confidence intervals.

\paragraph{Which design earns a scale pass.} Under pilot-scope evidence carried to the proposed scope by a transfer record that states only a value-side ratio (59 of 59 caught at 300 cases per day), no design earns a pass (Table~\ref{tab:qevpass}): each is insufficient or fails. If 59 live seeded attempts, none missed, are measured at the proposed scope, the following earn a pass on the risk check and the engine's value gate: per-item approval at a served volume of 600 per day (conditional on the value basis; see below), sampled review at 1,451 reviews per day, D2 at 1,400 cases per day, and D4 with a \$100 limit; D2 at the proposed scope earns one only if the automated check's recall is designed at 0.90 or more. With an observed 0 of 59 none passes. D2 at the proposed scope, the highest-value design, does not earn a pass.

\begin{table}[t]
\centering
\scriptsize
\caption{Which designs earn a scale pass under the evidence standard. Each cell is the quantified check / the engine decision (RA's other synthetic evidence held fixed). D1c is sampled review at the 1,451 reviews/day boundary; D2* is D2 with an assumed automated-check recall of 0.90 (above the 0.8824 break-even). The last column is the smallest number of seeded attempts, none missed, that can support the design's required containment (blank when the stated design fails by itself). $^\dagger$ Conditional: the engine's value gate uses RA's value basis (fixed 600,000 cases per year, \$150,000 run cost) and ignores review cost and served volume; Table~\ref{tab:qdesigns} gives a net value of $-$\$21,900 for this design on the RF cost basis. See the text below.}
\label{tab:qevpass}
\begin{tabularx}{\textwidth}{lXXXXl}
\toprule
\textbf{Design} & \textbf{Scope} & \textbf{E0: pilot-scope evidence, value-side transfer record} & \textbf{E1: matched scope, 59/59} & \textbf{E2: matched scope, 0/59} & \textbf{Zero-miss attempts needed} \\
\midrule
D0 & proposed scope & fail / Redesign & fail / Redesign & fail / Redesign & -- \\
D1a & proposed scope & insufficient (no risk-side rule) / Extend pilot & fail / Redesign & fail / Redesign & -- \\
D1b & proposed scope & insufficient (no risk-side rule) / Extend pilot & fail (queue) / Redesign & fail (queue) / Redesign & -- \\
D1b & scope change: served volume 600/day & insufficient (no risk-side rule) / Extend pilot & pass / Scale\textsuperscript{$\dagger$} & fail / Redesign & -- \\
D1c & proposed scope & insufficient (no risk-side rule) / Extend pilot & pass / Scale & fail / Redesign & 19 \\
D2 & proposed scope & insufficient (no risk-side rule) / Extend pilot & fail / Redesign & fail / Redesign & -- \\
D2 & scope change: volume 1,400/day & insufficient (no risk-side rule) / Extend pilot & pass / Scale & fail / Redesign & 9 \\
D3 & proposed scope & insufficient (no risk-side rule) / Extend pilot & fail / Redesign & fail / Redesign & -- \\
D4 & scope change: limit \$100 (loss per escape \$100 assumed) & insufficient (no risk-side rule) / Extend pilot & pass / Scale & fail / Redesign & 5 \\
D2* & proposed scope & insufficient (no risk-side rule) / Extend pilot & pass / Scale & fail / Redesign & 11 \\
\bottomrule
\end{tabularx}
\end{table}

\paragraph{Reading the D1b row (Table~\ref{tab:qdesigns} against Table~\ref{tab:qevpass}).} Per-item approval with served volume restricted to 600 cases per day is a scale pass in Table~\ref{tab:qevpass} (matched evidence, 59 of 59) but has a net value of $-$\$21{,}900 per year in Table~\ref{tab:qdesigns}. The two figures answer different questions on different cost bases. Table~\ref{tab:qdesigns} uses the scenario RF economics (a \$4 saving per case, kept on 50\% of reviewed cases, so \$2.00; a \$1.80 review cost; a residual loss of \$0.30 per served case), which gives $438{,}000 - 394{,}200 - 65{,}700 = -21{,}900$. The engine's value gate instead annualizes RA's measured pilot effect at a fixed 600,000 cases per year against a fixed \$150,000 run cost, whatever control is chosen: it includes neither the review cost nor the served volume. The engine label ``SCALE'' for D1b at 600 per day is therefore conditional on that value basis, not a recommendation to deploy. In RF terms each served case nets $2.00 - 1.80 - 0.30 = -\$0.10$ at any queue capacity, so raising throughput does not change the sign (cost and value both scale with volume). The sign changes only if the review cost falls below \$1.70 (net zero; \$1.50 gives +\$43,800), or the share of savings kept on reviewed cases exceeds 52.5\%. Before recommending this design, the two value estimates must be reconciled on one scope and one cost basis; until then the pass is a statement about the risk check and RA's value gate, and the value is not shown to be positive.

\subsection{What seeded attempts can establish}
Seeded attempts establish containment for the seeded mix. They do not establish prevalence, representativeness, the false-block rate on legitimate traffic or strategic behavior change; natural traffic establishes those and cannot establish containment of rare attempt types it has not yet seen. With exact one-sided bounds at 95\% confidence, 29 attempts with none missed show containment of at least 90\% and 59 show at least 95\% (28 and 58 do not); Table~\ref{tab:qevid} gives the counts with up to three misses. Expected traffic differs from the traffic needed to \emph{observe} enough attempts with a chosen probability: at 1\% prevalence, 59 attempts are 5,900 actions in expectation but need 7,211 actions for a 95\% chance of seeing at least 59. Simultaneous bounds on $q$, $d$ and an assumed bound on $l$ are combined by Bonferroni; the bound on $l$ is an assumption and is labelled so, and conservative input bounds and tail scenarios are reported separately. Tail measures such as conditional value at risk~\cite{rockafellar2000optimization} would be a policy extension, not a gate here. EVPI is only a ceiling: in a stated model (prior containment Beta(14,6), a \$1.5M operating gain, expansion only if the gate passes on the posterior mean, 59 seeded attempts) EVPI is \$1,139,964 and EVSI for that specific experiment is \$364,648. Both depend on those synthetic assumptions.

\begin{table}[t]
\centering
\footnotesize
\caption{Seeded attempts needed to show containment at one-sided 95\% confidence (exact binomial), and the natural traffic needed to see that many unsafe attempts at 1\% prevalence. Binomial and negative-binomial calculations agree.}
\label{tab:qevid}
\begin{tabularx}{\textwidth}{lXXXXXX}
\toprule
\textbf{Containment to show} & \textbf{Attempts, 0 missed} & \textbf{1 missed} & \textbf{2 missed} & \textbf{3 missed} & \textbf{Expected traffic at q = 1\%} & \textbf{Traffic for a 95\% chance of enough attempts} \\
\midrule
90\% & 29 & 46 & 61 & 76 & 2,900 & 3,834 \\
95\% & 59 & 93 & 124 & 153 & 5,900 & 7,211 \\
\bottomrule
\end{tabularx}
\end{table}

\section{Sources and status of rules}\label{sec:sources}
We separate what comes from an authoritative source from what is PilotGate prototype policy. The R-BED and V-BED presentation slide supplied by the author gives the \emph{presentation-level} structure: the R1--R4 risk flow and the V1--V4 / E1--E4 value stages. We attempted to open the October 2026 R-BED release and crosswalk but could not access it, so \textbf{no section or version of that release is cited}. Earlier author drafts of a multiplicative combination of three value factors disagree about averaging versus maximum, and the slide does not define one; we therefore do \emph{not} attribute any product of V1, V2 and V3 to R-BED and use no ordinal product. Everything below is a PilotGate policy assumption unless a source is named: the cell values of Table~\ref{tab:risktable}, the R1 adjustment, the cut points on the measured error rate, the appetite tiers, evidence level required by projected value size, the durability thresholds (study size, duration, 75\% stability share), the transport caps and haircut, the control-test age limit, and the funding hurdle and horizon, and, for the quantified check of Section~\ref{sec:quant}, the appetite, the scenario inputs and the evidence standard. The deterrence threshold of that section is derived directly from the stated payoffs, with the economic model of crime~\cite{becker1968crime} as context. All are configurable in \texttt{fixtures/policy.json}. Related work informs only general design: GRADE and evidence-standards ladders for graded evidence~\cite{guyatt2008grade,sherman1997preventing,puttick2013standards}; stochastic acceptability analysis and portfolio decision analysis for ranking under uncertainty~\cite{lahdelma1998smaa,salo2011portfolio}; value-of-information theory~\cite{raiffa1961applied}; exact binomial and rule-of-three bounds for error rates~\cite{clopper1934use,hanley1983zero}; Welch's unequal-variance comparison~\cite{welch1947generalization}; the NIST AI RMF and the EU AI Act for risk-management context~\cite{nist2023airmf,euaiact2024}; and use-case prioritization from practice~\cite{appliedai2020prioritize,bodendorf2025usecase}.

\section{Results on the running example}\label{sec:results}
\subsection{Screening and funding}
Table~\ref{tab:example} gives the pilot rung and residual risk. RA and RB reach a human-approved live pilot; RC is limited to shadow mode, and D1 (a vendor model on cardholder data with no lawful basis in the synthetic fixture) is held by a hard blocker. Table~\ref{tab:fund} shows funding, and its column ``EVSI belongs to'' says which experiment each EVSI figure describes. For RA, K1 and P1 it is a live pilot that the screen permits, and EVSI exceeds cost. \textbf{RC.} The \$86,105 EVSI is the value of the designed \$200,000 live pilot, which the pilot screen does not permit (RC is limited to shadow mode); RC's EVPI bound (\$88,629) is below that cost, so the live design is not recommended. The recommended \$15,000 shadow run measures no realized value, so its EVSI on value is zero. What it can inform is a risk decision: whether an error-rate bound at the shadow scope could later support the pilot screen for an approval-gated live pilot. We do not model the value of that risk information, and the run has no independent value rationale. \textbf{RB.} The \$335,104 EVSI (EVPI \$341,766, cost \$140,000) is the value of the modeled decision to scale at the proposed scope (\$200 limit, 1,644 cases per day). That decision cannot reach SCALE there: even with every required control credited, RB's best-case risk tier (CRITICAL) exceeds the HIGH appetite, so the structural risk rule fails the risk gate whatever the pilot measures (Section~\ref{sec:risk}, Table~\ref{tab:scale}). The figure is therefore not a funding justification for earning approval at the proposed scope. The recommendation for RB is therefore to \textbf{redesign the scope, then reassess pilot funding}; the code's output for the current scope is the label REDESIGN\_SCOPE\_THEN\_REASSESS, not a pilot approval. The \$140,000 pilot (a \$50 limit, 150 cases per day) could inform a redesigned lower-scope decision, but we do not compute an EVSI for a lower scope, because the fixture states no value drivers, scope or pilot design for one. Table~\ref{tab:plan} states what each experiment can teach and the decision it could change.

\begin{table}[t]
\centering
\footnotesize
\caption{The running example at three operating scopes: refund recommender (RA), approval-gated executor (RB) and autonomous capped executor (RC). Residual risk is the R4 level with its R1/R2/R3 inputs (prototype policy, Section~\ref{sec:risk}).}
\label{tab:example}
\begin{tabularx}{\textwidth}{lllX}
\toprule
\textbf{Use case} & \textbf{Pilot rung} & \textbf{Residual risk, pilot scope} & \textbf{Residual risk, target scope} \\
\midrule
RA & HUMAN\_\allowbreak{}APPROVED\_\allowbreak{}LIVE & LOW [R1 2 | R2 1 | R3 2] & HIGH [R1 2 | R2 2 | R3 3] \\
RB & HUMAN\_\allowbreak{}APPROVED\_\allowbreak{}LIVE & HIGH [R1 2 | R2 2 | R3 3] & CRITICAL [R1 2 | R2 3 | R3 3] \\
RC & SHADOW & HIGH [R1 2 | R2 2 | R3 3] & CRITICAL [R1 2 | R2 3 | R3 3] \\
\bottomrule
\end{tabularx}
\end{table}

\begin{table}[t]
\centering
\footnotesize
\caption{Funding decision. EVPI is an upper bound on what perfect information could be worth. EVSI is for the specific designed experiment named in the column ``EVSI belongs to''. RC's EVSI is for the \$200,000 live design, which the pilot screen does not permit; the recommended \$15,000 shadow run measures no realized value, so its value-side EVSI is zero, and it informs a risk decision (whether a later approval-gated pilot may pass the screen). RB's EVSI is the value of the modeled scale decision at the proposed scope, which cannot reach SCALE there (Section~\ref{sec:results}); it is not a funding justification at that scope. Rows follow the EVSI ranking, which does not account for RB's structural risk finding; that is why RB is first and yet not recommended for a pilot at its proposed scope.}
\label{tab:fund}
\begin{tabularx}{\textwidth}{llll>{\raggedright\arraybackslash}p{3.4cm}>{\raggedright\arraybackslash}p{1.9cm}X}
\toprule
\textbf{Use case} & \textbf{Cost} & \textbf{EVPI bound} & \textbf{EVSI} & \textbf{EVSI belongs to} & \textbf{Information case} & \textbf{Recommendation} \\
\midrule
RB & \$140,000 & \$341,766 & \$335,104 & live pilot; the modelled scale decision cannot reach SCALE at the proposed scope & supported & redesign the scope, then reassess pilot funding \\
K1 & \$70,000 & \$227,354 & \$226,455 & live pilot, permitted & supported & fund live pilot \\
RA & \$80,000 & \$157,211 & \$151,051 & live pilot, permitted & supported & fund live pilot \\
RC & \$200,000 & \$88,629 & \$86,105 & hypothetical live design; the screen does not permit it & EVPI below cost & shadow run only \\
P1 & \$60,000 & \$111,678 & \$110,154 & live pilot, permitted & supported & fund live pilot \\
D1 & \$90,000 & n/a & n/a &  & none & not eligible \\
RC shadow run & \$15,000 & - & \$0 & shadow run: no value measured; informs a risk decision & risk information only & shadow run only \\
\bottomrule
\end{tabularx}
\end{table}

\begin{table}[t]
\centering
\footnotesize
\caption{What each funded experiment can teach and the decision it could change.}
\label{tab:plan}
\begin{tabularx}{\textwidth}{lXXX}
\toprule
\textbf{Use case} & \textbf{Experiment} & \textbf{Decision that could change} & \textbf{Independent rationale} \\
\midrule
RA & live pilot at the pilot scope (3000 cases per arm, 8 weeks; a concurrent control is needed & scale versus extend, redesign or stop at the proposed scope (the value gate) & direct operating benefit during the pilot (assumption) \\
RB & live pilot at the pilot scope (2000 cases per arm, 8 weeks; a concurrent control is needed & a redesigned, lower-scope expansion decision. The pilot cannot earn approval at the proposed scope: best-case risk tier CRITICAL exceeds the HIGH appetite even with every required control credited & direct operating benefit during the pilot (assumption) \\
RC & shadow run on live traffic: the AI decides, nothing is executed & the pilot screen for the next rung (whether a live approval-gated pilot may be authorized), not the scale decision & none stated \\
P1 & live pilot at the pilot scope (600 cases per arm, 6 weeks; a concurrent control is needed  & scale versus extend, redesign or stop at the proposed scope (the value gate) & direct operating benefit during the pilot (assumption); Regulatory exam finding on reconci \\
K1 & live pilot at the pilot scope (6000 cases per arm, 8 weeks; a concurrent control is needed & scale versus extend, redesign or stop at the proposed scope (the value gate) & direct operating benefit during the pilot (assumption) \\
\bottomrule
\end{tabularx}
\end{table}

\subsection{Expansion decisions}
Table~\ref{tab:scale} shows the expansion decisions. RA passes both gates. RB is INSUFFICIENT on value (its pilot effect does not apply to the proposed scope) and fails risk. K1 shows that strong capability and adoption do not guarantee value: its incremental net value is negative and below the hurdle even at the optimistic end, so the value gate fails. P1 has only E2 evidence where E3 is required, so it extends its pilot. RC has only shadow evidence (E1).

\begin{table}[t]
\centering
\footnotesize
\caption{Expansion decisions at the proposed scope. Annual value is the interval of incremental net value per year; Req.\ is the evidence level the projected value requires.}
\label{tab:scale}
\begin{tabularx}{\textwidth}{lllllllX}
\toprule
\textbf{Use case} & \textbf{Design} & \textbf{Evid.} & \textbf{Req.} & \textbf{Value} & \textbf{Risk} & \textbf{Decision} & \textbf{Annual value (low / base / high)} \\
\midrule
RA & randomized & E4 & E3 & PASS & PASS & SCALE & \$220,294 / \$465,234 / \$710,174 \\
RB & randomized & E4 & E3 & INSUFFICIENT & FAIL & REDESIGN & \$118,887 / \$255,999 / \$393,111 \\
K1 & randomized & E4 & E3 & FAIL & INSUFFICIENT & REDESIGN & \$-224,017 / \$-161,789 / \$-99,561 \\
P1 & pre\_\allowbreak{}post & E2 & E3 & INSUFFICIENT & INSUFFICIENT & EXTEND\_\allowbreak{}PILOT & \$-235,792 / \$-107,026 / \$21,740 \\
RC & shadow & E1 & E4 & INSUFFICIENT & FAIL & REDESIGN & n/a / n/a / n/a \\
\bottomrule
\end{tabularx}
\end{table}

\begin{table}[t]
\centering
\footnotesize
\caption{Implementation status mapped to executive action, with whether a separately evaluated lower-scope experiment remains open.}
\label{tab:exec}
\begin{tabularx}{\textwidth}{lllX}
\toprule
\textbf{Use case} & \textbf{Status} & \textbf{Executive action} & \textbf{Lower-scope experiment?} \\
\midrule
RA & SCALE & Deploy/Scale & - \\
RB & REDESIGN & Redesign & a redesigned (lower) scope is a new candidate and is screened afresh \\
K1 & REDESIGN & Redesign & a redesigned (lower) scope is a new candidate and is screened afresh \\
P1 & EXTEND\_\allowbreak{}PILOT & Extend pilot with constraints & the extended pilot is the lower-scope experiment \\
RC & REDESIGN & Redesign & a redesigned (lower) scope is a new candidate and is screened afresh \\
\bottomrule
\end{tabularx}
\end{table}

\subsection{Before and after}
Table~\ref{tab:beforeafter} compares expansion decisions with the 153-test revision. The software was changed (R1--R4 decision table, stage-graded evidence, mechanism model, executive actions, matched-scope case), yet no pilot-screening or expansion decision changed, and one funding recommendation did: RB moved from FUND\_PILOT to ``redesign the scope, then reassess pilot funding'' once the structural risk finding was applied to its funding case (Table~\ref{tab:fund}; see the companion changed-decision table). Table~\ref{tab:regress} gives the mutation regressions: adding a blocker, making the scope unknown, removing guardrail results, and weakening mediation each stop the expansion. Table~\ref{tab:scope} shows the scope exercise: raising volume tenfold, raising or removing the limit, sampling review, or raising authority each force reverification.

\begin{table}[t]
\centering
\footnotesize
\caption{Expansion decisions before and after replacing the product-style rating with the R1--R4 table and stage-graded evidence. No headline decision changed; the explanations and evidence ladder did.}
\label{tab:beforeafter}
\begin{tabularx}{\textwidth}{lXXl}
\toprule
\textbf{Use case} & \textbf{153-test revision} & \textbf{This revision} & \textbf{Changed} \\
\midrule
RA & SCALE (value PASS, risk PASS) & SCALE (value PASS, risk PASS) & no \\
RB & REDESIGN (value INSUFFICIENT, risk FAIL) & REDESIGN (value INSUFFICIENT, risk FAIL) & no \\
K1 & REDESIGN (value FAIL, risk INSUFFICIENT) & REDESIGN (value FAIL, risk INSUFFICIENT) & no \\
P1 & EXTEND\_\allowbreak{}PILOT (value INSUFFICIENT, risk INSUFFICIENT) & EXTEND\_\allowbreak{}PILOT (value INSUFFICIENT, risk INSUFFICIENT) & no \\
RC & REDESIGN (value INSUFFICIENT, risk FAIL) & REDESIGN (value INSUFFICIENT, risk FAIL) & no \\
\bottomrule
\end{tabularx}
\end{table}

\begin{table}[t]
\centering
\footnotesize
\caption{Regression mutations of the passing RA scale case. Each unsafe change is stopped, deferred, or reverted to an extended pilot or redesign.}
\label{tab:regress}
\begin{tabularx}{\textwidth}{XXXl}
\toprule
\textbf{Mutation of the passing case} & \textbf{Before} & \textbf{After} & \textbf{Executive action} \\
\midrule
none (control) & SCALE (value PASS, risk PASS) & SCALE (value PASS, risk PASS) & Deploy/Scale \\
1. add a hard blocker & SCALE (value PASS, risk PASS) & BLOCKED (value PASS, risk PASS) & Stop \\
2. operating scope unknown & SCALE (value PASS, risk PASS) & DEFER (value PASS, risk PASS) & none (deferred) \\
3. remove all guardrail results & SCALE (value PASS, risk PASS) & EXTEND\_\allowbreak{}PILOT (value INSUFFICIENT, risk PASS) & Extend pilot with constraints \\
4. weaken mediation per-item to sampled, reuse evidence & SCALE (value PASS, risk PASS) & REDESIGN (value INSUFFICIENT, risk FAIL) & Redesign \\
\bottomrule
\end{tabularx}
\end{table}

\begin{table}[t]
\centering
\footnotesize
\caption{Scope exercise: each change to the evaluated scope forces re-verification (full reasons in the companion tables).}
\label{tab:scope}
\begin{tabularx}{\textwidth}{Xlll}
\toprule
\textbf{Proposed scope change to RA} & \textbf{Decision} & \textbf{Value} & \textbf{Risk} \\
\midrule
as evaluated (limit \$500, 1,644/day, per-item) & SCALE & PASS & PASS \\
volume x10 (16,440/day) & EXTEND\_\allowbreak{}PILOT & INSUFFICIENT & INSUFFICIENT \\
limit \$1,000 & EXTEND\_\allowbreak{}PILOT & INSUFFICIENT & INSUFFICIENT \\
uncapped & EXTEND\_\allowbreak{}PILOT & INSUFFICIENT & INSUFFICIENT \\
sampled review & REDESIGN & INSUFFICIENT & FAIL \\
execute with approval & REDESIGN & INSUFFICIENT & FAIL \\
\bottomrule
\end{tabularx}
\end{table}

\subsection{Matched-scope scale case and transferability}\label{sec:matched}
The matched-scope case is a \emph{synthetic matched-scope fixture that verifies the decision logic}. Its evidence scope (volume and control-test scope) is set to the proposed 1,644 cases/day and a fresh synthetic sample is drawn; we do not simulate operations or control behavior at that throughput. Because the evidence scope coincides with the proposal, no transfer is needed and both gates pass. The economic result is simulated E4 evidence at the evidence scope, with an assumed run cost, and the annual figure remains a forecast. The transferred case instead takes evidence measured at 300 cases/day and relies on a stated volume-transfer assumption; Table~\ref{tab:matched} contrasts them. Table~\ref{tab:stress} stresses that assumption. Removing it, or stating it without evaluating it, gives EXTEND\_PILOT. The decision holds for effect haircuts of 1.00 to 0.70 and flips at 0.50. A stricter value threshold of \$150,000 still passes; \$200,000 does not. The transfer rests on eight weekly observations and should be treated as limited evidence, not a validated assumption. The two gates have different transfer ranges. The risk-side record covers 300 to 1,644 cases per day, the proposed scope. The value-side record for RA covers 300 to 1,700 cases per day, with a 0.85 haircut and a maximum volume ratio of 6.0. Between 1,645 and 1,700 cases per day, with pilot-scope containment evidence only, the value gate can rely on a covered transfer while the risk gate is INSUFFICIENT because the control evidence does not apply; the decision is then EXTEND\_PILOT. Above 1,700 neither gate is covered.

\begin{table}[t]
\centering
\footnotesize
\caption{Matched-scope versus transferred evidence for RA. The matched case is a fixture relabelling the evidence scope; its economics are simulated E4 and the annual figure is a forecast.}
\label{tab:matched}
\begin{tabularx}{\textwidth}{Xlll}
\toprule
\textbf{Case} & \textbf{Decision} & \textbf{Relies on transfer} & \textbf{Evidence scope} \\
\midrule
matched scope (evidence at 1,644/day) & SCALE & no & 1,644/day \\
transferred (evidence at 300/day, proposal 1,644/day) & SCALE & yes & 300/day \\
\bottomrule
\end{tabularx}
\end{table}

\begin{table}[t]
\centering
\footnotesize
\caption{Transferability stress test of the 300 to 1,644 cases/day expansion. The transfer rests on eight weekly observations and is limited evidence.}
\label{tab:stress}
\begin{tabularx}{\textwidth}{Xlll}
\toprule
\textbf{Variant} & \textbf{Decision} & \textbf{Value gate} & \textbf{Low end (\$/yr)} \\
\midrule
as evaluated (volume transport, haircut 0.85, hurdle \$100,000) & SCALE & PASS & 164750.0 \\
transfer assumption removed & EXTEND\_\allowbreak{}PILOT & INSUFFICIENT &  \\
assumption stated but not evaluated & EXTEND\_\allowbreak{}PILOT & INSUFFICIENT &  \\
effect haircut 1.00 & SCALE & PASS & 220294.0 \\
effect haircut 0.85 & SCALE & PASS & 164750.0 \\
effect haircut 0.70 & SCALE & PASS & 109206.0 \\
effect haircut 0.50 & EXTEND\_\allowbreak{}PILOT & INSUFFICIENT & 35147.0 \\
effect haircut 0.30 & EXTEND\_\allowbreak{}PILOT & INSUFFICIENT & -38912.0 \\
stricter value threshold \$150,000 & SCALE & PASS & 164750.0 \\
stricter value threshold \$200,000 & EXTEND\_\allowbreak{}PILOT & INSUFFICIENT & 164750.0 \\
stricter value threshold \$250,000 & EXTEND\_\allowbreak{}PILOT & INSUFFICIENT & 164750.0 \\
\bottomrule
\end{tabularx}
\end{table}

\section{Evaluation example: a public AWS sample}\label{sec:aws}
To show how a single-pass score differs, we reproduce the prioritization in a public AWS sample~\cite{awssamples} on a 26-criterion synthetic fixture. As implemented it returns a composite of 4.4, a risk proxy of 25 out of 25 and readiness 5.0, and the verdict is CONDITIONAL GO, because its NO-GO branch needs all three conditions together. Its documentation describes any one condition as sufficient, which gives NO GO. We compare both with the comment-derived OR variant and do not treat either as an authorization (Table~\ref{tab:aws}). Table~\ref{tab:threeway} places these beside the framework: for RC the OR rule says NO GO while the framework gives a shadow pilot at intake and REDESIGN at the larger scope. The point is not that one system is right, but that a use-case score does not say at which scope the use case is acceptable.

\begin{table}[t]
\centering
\footnotesize
\caption{Reproduction of the AWS sample prioritization on the 26-criterion fixture. Neither verdict is an authorization.}
\label{tab:aws}
\begin{tabularx}{\textwidth}{Xllll}
\toprule
\textbf{Variant} & \textbf{Composite} & \textbf{Risk} & \textbf{Readiness} & \textbf{Verdict} \\
\midrule
AWS as implemented (NO GO needs all three) & 4.4 & 25 & 5.0 & CONDITIONAL GO \\
AWS with documented OR rule (any one) & 4.4 & 25 & 5.0 & NO GO \\
\bottomrule
\end{tabularx}
\end{table}

\begin{table}[t]
\centering
\footnotesize
\caption{Three-way comparison on the synthetic use cases: AWS as implemented, AWS under the documented OR rule, and this framework at the proposed scope.}
\label{tab:threeway}
\begin{tabularx}{\textwidth}{l>{\raggedright\arraybackslash}p{2.2cm}llllX}
\toprule
\textbf{Use case} & \textbf{State} & \textbf{Composite} & \textbf{Risk} & \textbf{AWS as implemented} & \textbf{AWS OR rule} & \textbf{This framework} \\
\midrule
RA & intake & 3.38 & 9 & CONDITIONAL GO & CONDITIONAL GO & HUMAN\_\allowbreak{}APPROVED\_\allowbreak{}LIVE \\
RA & post\_\allowbreak{}pilot & 3.56 & 6 & GO & GO & SCALE \\
RB & intake & 2.9825 & 12 & CONDITIONAL GO & CONDITIONAL GO & HUMAN\_\allowbreak{}APPROVED\_\allowbreak{}LIVE \\
RB & post\_\allowbreak{}pilot & 3.0575 & 12 & CONDITIONAL GO & CONDITIONAL GO & REDESIGN \\
RC & intake & 3.045 & 25 & CONDITIONAL GO & NO GO & SHADOW \\
RC & post\_\allowbreak{}pilot & 3.075 & 25 & CONDITIONAL GO & NO GO & REDESIGN \\
P1 & intake & 3.27 & 4 & CONDITIONAL GO & CONDITIONAL GO & HUMAN\_\allowbreak{}APPROVED\_\allowbreak{}LIVE \\
P1 & post\_\allowbreak{}pilot & 3.105 & 16 & CONDITIONAL GO & CONDITIONAL GO & EXTEND\_\allowbreak{}PILOT \\
K1 & intake & 3.2675 & 4 & CONDITIONAL GO & CONDITIONAL GO & HUMAN\_\allowbreak{}APPROVED\_\allowbreak{}LIVE \\
K1 & post\_\allowbreak{}pilot & 3.2525 & 9 & CONDITIONAL GO & CONDITIONAL GO & REDESIGN \\
D1 & intake & 2.58 & 15 & CONDITIONAL GO & CONDITIONAL GO & HOLD \\
\bottomrule
\end{tabularx}
\end{table}

\section{Verification, scope of claims and limitations}\label{sec:verify}
\paragraph{Federal intake.} A pinned copy of the 2025 federal inventory~\cite{federalinventory2025} has 3,611 rows and 2,901 unique identifiers; 152 are in pre-deployment or pilot stage. A deterministic stratified selection of 12 records (Table~\ref{tab:fed}) is used for \emph{intake triage only}: agency-reported benefits are treated as intake hypotheses with no evidence level, and rankings driven by proxy fields are exploratory.
\begin{table}[t]
\centering
\footnotesize
\caption{Twelve deterministically selected inventory records used for intake triage only. Reported benefits are hypotheses with no evidence level.}
\label{tab:fed}
\begin{tabularx}{\textwidth}{lXlll}
\toprule
\textbf{Inventory key} & \textbf{Use case} & \textbf{Benefit claim} & \textbf{Pilot rung} & \textbf{Median rank} \\
\midrule
VA|VA-25-5578 & AICES & qualitative & SIMULATION & 11.0 \\
VA|VA-25-3732 & Modeling to Learn (MTL) & qualitative & SIMULATION & 11.0 \\
VA|VA-25-3480 & Smart Pension Automation & quantified & SIMULATION & 8.0 \\
SBA|SBA-30 & SBA.gov External Customer Chatbot & quantified & HOLD & 3.0 \\
SBA|SBA-35 & AI Incident Summarization & quantified & HOLD & 3.0 \\
SBA|SBA-13 & Generative AI for SBA.gov product training & qualitative & HOLD & 5.0 \\
VA|VA-25-1184 & Ask VA Inquiry Automated Category Classification & quantified & SIMULATION & 8.0 \\
NASA|NASA-744 & Enhancements to T-Rex and D-Rex & quantified & SIMULATION & 8.0 \\
NASA|NASA-743 & Ask Tech Port & quantified & SIMULATION & 8.0 \\
TREAS|REQ-2025-02087 & AI-assisted Procurement POC & qualitative & SIMULATION & 6.0 \\
EPA|27 & AI Assistants in Esri Tools & qualitative & SIMULATION & 2.0 \\
SBA|SBA-23 & Email, Document, and Composing AI Editor & qualitative & SIMULATION & 2.0 \\
\bottomrule
\end{tabularx}
\end{table}

\paragraph{What the evidence is.} We distinguish three classes of claim.
\begin{itemize}\setlength\itemsep{0pt}
\item \textbf{Software verification.} The package passes 279 tests with the pinned federal inventory present; without it 275 pass and 4 are skipped, and skipped tests are reported as skipped, not passed (153 in the baseline revision, reproduced before any edit). They cover the decision table, precedence, evidence applicability, mechanisms, executive actions, funding, regressions, and the quantified check, including three regressions in which an unsupported control credit had returned SCALE (Appendix~\ref{app:evidence}).
\item \textbf{Synthetic demonstrations.} All refund, reconciliation, knowledge-assistant and dispute cases, the assumed 5\% run-cost drift (so E4 economics are simulated), and the control tests are synthetic fixtures. They show that the logic behaves as designed, not that any organization's numbers look like these.
\item \textbf{Claims requiring field validation.} That the R1--R4 cell values, appetite tiers, thresholds and transfer assumptions suit a real organization, and that the stage ladder predicts realized value, are untested here.
\end{itemize}
\paragraph{Limitations.} The risk table and appetite are prototype policy; the October 2026 R-BED release was not accessible; the transfer assumption rests on eight weekly points; the matched-scope case uses a re-scoped control-test set; the quantified check treats the attempt rate and the loss per escape as stated inputs and every scenario parameter is synthetic; the \$100-limit result rests on an assumed loss per escape; the pessimistic annualization corner of RA sits at or below the hurdle, and whether that corner should gate SCALE is a policy question we leave open; and funding is deliberately conservative about learning. The framework supports a funding and scoping discussion; it is not a compliance determination.

\paragraph{Fixtures behind the tables.} Two synthetic RA pilots are used, and both return SCALE. The canonical pilot (\texttt{synth.build\_all()}, 3,044 treated cases, ITT effect ratios 0.6172, 1.0254 and 1.4336) generates the expansion, stage, regression, scope, stress and matched tables (Tables~\ref{tab:scale}, \ref{tab:exec}, \ref{tab:beforeafter}, \ref{tab:regress}, \ref{tab:scope}, \ref{tab:stress} and, against a freshly drawn matched sample, Table~\ref{tab:matched}). The test-helper pilot (3,000 cases per arm, ratios 0.60, 1.00 and 1.40) generates the quantified-engine tables (Tables~\ref{tab:qengine}, \ref{tab:qevpass} and \ref{tab:qregress}). The RF design comparison, the worked scenario and the sensitivity tables are closed-form and use no pilot. The fixtures differ only in the low effect ratio, so the gate decision agrees while the extrapolated low end differs: $0.85\times 0.60\times 600{,}000 - 150{,}000 = \$156{,}000$ for the test-helper pilot and \$164,750 for the canonical pilot (Table~\ref{tab:stress}, baseline row).

\paragraph{Reproducibility.} The companion package regenerates every table and figure with \texttt{python reproduce.py} and this source, including Figures 3 and 4, with \texttt{python paper/build\_paper.py}, which fails if prose diverges from results. Pinned inputs are AWS revision \texttt{7a0f137}, federal inventory commit \texttt{06c7ebe}, and fixture hashes in \texttt{results/environment.json}.

\section*{Author information and disclosures}
\paragraph{Author.} Vishal Srivastava, Whiting School of Engineering, Johns Hopkins University, Baltimore, MD 21218, USA (ORCID \href{https://orcid.org/0000-0002-5651-8365}{0000-0002-5651-8365}; \texttt{vsrivas7@jhu.edu}). R-BED and V-BED are frameworks developed by the author; the rules built on them here are PilotGate policy unless a source is named (Section~\ref{sec:sources}).
\paragraph{Funding.} This research received no external funding.
\paragraph{Competing interests.} The author declares no competing interests.
\paragraph{Use of AI tools.} The author used Claude (Anthropic) to assist with drafting, code, tests and verification of this paper and its companion package. The author reviewed the output and is responsible for the content.
\paragraph{Data and code.} All demonstrations use synthetic data, except the public inventories named in Section~\ref{sec:sources}. No human subjects were involved. The companion package regenerates every table and figure (see Reproducibility).

\bibliographystyle{plainnat}
\bibliography{references}

\appendix
\section{Derivations}\label{app:deriv}
\paragraph{Required containment and volume ceiling.} With $N=365V$, $L = Nq(1-d)l \le A$ gives $d \ge 1 - A/(365\,V q l)$, which is zero while $V \le A/(365\,q l) = 410.96$ cases per day (so no containment is required at 300) and $0.7500$ at 1,644. Under sampled review $d(V)=\rho\min\{1,B/V\}$. For $V \ge B$ the test $365\,q l\,(V-\rho B) \le A$ gives the single ceiling $V_{\max} = \rho B + A/(365\,q l) = 510 + 410.96 = 920.96$. The expected loss is continuous at $V=B$ and increasing in $V$, so there is one ceiling and the test fails above it.
\paragraph{Minimum capacity.} At $V = 1{,}644$ the requirement $\rho B / V \ge d_{\min}$ gives $B \ge d_{\min}V/\rho = 1{,}450.64$, so 1,451 is the smallest whole number of reviews per day.
\paragraph{Break-even recall.} For D2 the containment is $r\rho$ with automated-check recall $r$ (while flagged items fit within review capacity), so the expected loss equals the appetite at $r^* = d_{\min}/\rho = 0.8824$.
\paragraph{Exact bounds.} If $x$ of $n$ seeded attempts are caught, the one-sided lower bound on containment at level $\alpha$ is $\alpha^{1/n}$ when $x=n$ and the Clopper--Pearson beta quantile otherwise~\cite{clopper1934use}. Showing containment of at least $d_0$ with none missed needs $n \ge \lceil \ln\alpha/\ln d_0\rceil$ attempts: 29 for $d_0=0.90$ and 59 for $0.95$. With zero observed failures in general the same logic gives the rule-of-three style bound~\cite{hanley1983zero}. When two stages are credited, each bound uses $\alpha/2$ (Bonferroni); stage independence is never assumed, because the second stage is measured among the escapes of the first.

\section{The evidence standard and its status codes}\label{app:evidence}
Table~\ref{tab:qregress} lists the three bypasses found in an independent review of an earlier implementation and what they return under the standard of Section~\ref{sec:evidence}. In the first, a pipeline with first-stage containment 1 and no evidence passed; in the second, point assumptions overrode an observed 0 of 59; in the third, a transfer record's value-side effect factor 0.85 was ignored, so the gate used the unadjusted recall.
\begin{table}[t]
\centering
\footnotesize
\caption{Three bypasses found in independent review of an earlier implementation of the quantified check, each of which returned SCALE, and their outcomes under the evidence standard.}
\label{tab:qregress}
\begin{tabularx}{\textwidth}{lXXX}
\toprule
\textbf{} & \textbf{Case (RA at 1,644 cases/day)} & \textbf{Before the evidence standard} & \textbf{After} \\
\midrule
1 & pipeline d1 = 1, evidence = None & pass / Scale & insufficient (evidence not applicable) / Extend pilot \\
2 & 1,451 reviews/day, observed 0/59 caught & pass / Scale & fail / Redesign \\
3 & 1,451 reviews/day, pilot-scope 59/59, transport record effect factor 0.85 & pass / Scale & insufficient (no risk-side rule) / Extend pilot \\
\bottomrule
\end{tabularx}
\end{table}

The stable status codes and their effect in the decision path are:
\begin{itemize}\setlength\itemsep{0pt}\footnotesize
\item \texttt{QL\_PASS}: this check passes (it satisfies that check only).
\item \texttt{QL\_FAIL\_EXPECTED\_LOSS}, \texttt{QL\_FAIL\_CAPACITY\_QUEUE}: the risk gate fails, so REDESIGN.
\item \texttt{QL\_INSUFFICIENT\_NO\_SPEC}, \texttt{QL\_INSUFFICIENT\_NO\_APPETITE}: required inputs or the appetite are missing.
\item \texttt{QL\_INSUFFICIENT\_EVIDENCE\_NOT\_APPLICABLE}: the evidence does not cover the proposed scope.
\item \texttt{QL\_INSUFFICIENT\_TRANSPORT\_RISK\_RULE}: coverage rests on a transfer that states no risk-side recall ratio.
\item \texttt{QL\_INSUFFICIENT\_EVIDENCE\_STRENGTH}: the design passes, the evidence is too weak to show it.
\end{itemize}
Three layers must be kept apart. (i) The \emph{quantified check} returns one status code. It assesses whether the evidence applies to the proposed scope \emph{before} it assesses the design, so inapplicable evidence returns \texttt{QL\_INSUFFICIENT\_EVIDENCE\_NOT\_APPLICABLE} even where the stated design would fail. (ii) The \emph{risk gate} combines that code with the ordinal rating, the vetoes and the mandatory controls. A failed mandatory control, a hard veto, a best-case tier above appetite or a quantified FAIL makes the gate FAIL; only if none of these holds does a quantified INSUFFICIENT make the gate INSUFFICIENT. (iii) The \emph{final action} follows the decision precedence of Section~\ref{sec:framework}: a risk FAIL gives REDESIGN and an INSUFFICIENT gate gives EXTEND\_PILOT. A code therefore does not determine the action by itself. In the sampled-review design D1c (600 reviews per day, 1,644 cases per day, appetite \$300,000), the stated design's modeled loss is \$827,820 in both of the following cases. With 59 of 59 caught at the pilot scope and no applicable transfer, the check is \texttt{QL\_INSUFFICIENT\_EVIDENCE\_NOT\_APPLICABLE}, reports no credited loss, and the decision is EXTEND\_PILOT. With 59 of 59 caught at the matched scope the evidence applies, the credited loss still exceeds the appetite, the check is \texttt{QL\_FAIL\_EXPECTED\_LOSS} and the decision is REDESIGN. The statement that every INSUFFICIENT code yields EXTEND\_PILOT holds only when no FAIL condition applies elsewhere in the risk gate. This describes the implemented behavior; no rule was changed. A required gate accepts only the evidence basis; point values are for planning and figures, and a required gate configured on them raises an error.

\section{Synthetic inputs and accepted evidence}\label{app:inputs}
\begin{itemize}\setlength\itemsep{0pt}
\item \textbf{Synthetic scenario inputs} (approved for implementation and sensitivity analysis, not calibrations): $q$, $l$, $A$, $\rho$, $B$, the automated-check recall and flag rate, the share of savings kept on reviewed cases, the recovery rate, the effect of a limit on loss per escape, the EVSI prior and gain, and the deterrence payoffs.
\item \textbf{Evidence the required gate accepts}: observed, live seeded-attempt counts whose scope covers the proposed scope (directly, or through an evaluated transfer that states a risk-side ratio), credited at the exact lower bound and never above the design value. These bound recall, pipeline containment, detection, recovery and (with an evaluated payoff assumption) deterrence.
\item \textbf{Stated inputs that the gate does not test}: the attempt rate $q$ and the loss per escape $l$. Their values drive the result and need their own evidence in an application.
\end{itemize}

\end{document}